\documentclass{book}
\usepackage[ISSUE]{liltdraft}

\crop[]         % Uncomment for centered pages.
\usepackage[comma]{natbib}  % CSLI Pubs favored bibliography package.
\usepackage{chapterbib}     % Makes a section out of References

\usepackage{url,xspace,relsize}
\usepackage{omwmacros}
\usepackage[j]{mtg2e}

\usepackage{amsmath,amssymb,amsthm,array,booktabs,enumitem,graphicx,here,hhline,multirow,threeparttable,wrapfig}
\usepackage[hang]{footmisc}
\usepackage[utf8]{inputenc}
\usepackage[russian,english]{babel}
\usepackage{caption}
\usepackage[ruled,vlined]{algorithm2e}

\DeclareMathOperator*{\argmax}{arg\,max}
\DeclareMathOperator*{\maximize}{\textrm{maximize}}
\newcommand{\specialcell}[2][c]{%
\begin{tabular}[#1]{@{}c@{}}#2\end{tabular}}

\title       {Extending and Improving Wordnet via Unsupervised Word Embeddings}
\author      {Mikhail Khodak,
              Andrej Risteski,
              Christiane Fellbaum,
              Sanjeev Arora\affiliation{Princeton University}}
\begin{document}

This work presents an unsupervised approach for improving WordNet that builds upon recent advances in document and sense representation via distributional semantics.
We apply our methods to construct Wordnets in French and Russian, languages which both lack good manual constructions.\footnote{Available at \url{http://nlp.cs.princeton.edu/PAWN/}.}
These are evaluated on two new 600-word test sets for word-to-synset matching and found to improve greatly upon synset recall, outperforming the best automated Wordnets in F-score.
Our methods require very few linguistic resources, thus being applicable for Wordnet construction in low-resources languages, and may further be applied to sense clustering and other Wordnet improvements.

\section{Introduction}
\label{sec:intro}

Since the development of the Princeton WordNet (PWN) and its successful application to computational linguistics and information retrieval \citep{Fellbaum:98}, there have been many efforts to extend it to other languages and improve its synset relations and sense associations.
Doing this by hand is difficult and resource-intensive, making automated methods desirable.
However, these are often tailored to a specific language structure or depend heavily on resource availability, complicating application to many languages.
We develop an unsupervised approach for synset representation and word-sense induction and apply it to automated Wordnet construction for French and Russian.
The method requires only an unannotated corpus in the target language and machine translation (MT) between that language and English.

The basis of our work is the use of {\em word embeddings}: representations of words as vectors, typically real and low-dimensional \citep{Turney:10}. 
Although many vector representations of synsets have been proposed, most already depend on Wordnet, limiting their use for building it in new languages.
We instead represent translated synset information directly using recent work on document representations \citep{sentence}.
We also apply a method for linear algebraic word-sense induction (WSI) to develop a sense-clustering procedure that can be further used to improve Wordnet construction \citep{polysemy}.

Our further contribution is the application of these representations to the {\em extend} approach for automated Wordnet construction \citep{Vossen:98}.
This framework assumes that synset relations are invariant across languages and generates a set of candidate synsets for each word $w$ in the target language by using a set of English translations of $w$  to query PWN (we refer to this as MT+PWN).
As the number of candidate synsets produced may be quite large, we need to select those synsets that are its appropriate senses.
Here a simple word embedding approach is to use a cutoff on the average similarity between a word and the synset's lemmas.
We find that using our synset representations improve greatly upon this baseline and outperforms other language-independent methods as well as language-specific approaches such as WOLF, the French Wordnet used by the Natural Language ToolKit \citep{wolf,omw,nltk}.

A further contribution is two 600-word test sets in French and Russian that are the largest and most comprehensive available, containing 200 each of nouns, verbs, and adjectives.
We construct them by presenting native speakers with all candidate synsets produced by MT+PWN and treating the senses picked as ``ground truth" for measuring accuracy.
Besides its size, our data sets also have the advantage of being separated by part-of-speech (POS), making evident differences in performance across POS.
With these test sets, we hope to address the difficulties in evaluating non-English Wordnets from the use of different and unreported data, incompatible metrics (e.g. matching synsets vs. retrieving synset lemmas), and differing cross-lingual dictionaries.

\section{Related Work}
\label{sec:related}

Much past work on automated Wordnets has focused on language-specific approaches --- using resources or properties specific to a language or language family.
Efforts for Korean \citep{Lee:00}, French \citep{wolf,wonef}, and Persian \citep{Montazery:10}, have found success in using bilingual corpora, expert knowledge, or Wordnets in related languages on top of an MT+PWN step. 
We compare to the Wordnet Libre du Fran\c{c}ais (WOLF), which leverages multiple European Wordnets \citep{wolf};
in our evaluation an embedding method outperforms the approach in F-score while having far fewer resource requirements.
Wordnet du Fran\c{c}ais (WoNeF), an extension of WOLF that combined linguistic models via a voting scheme \citep{wonef}, was found to have performance generally below WOLF's, so we compare to the earlier database.

There have also been recent vector approaches for Wordnet construction, specifically for an Arabic Wordnet \citep{AlTarouti:16} and a Bengali Wordnet \citep{Nasiruddin:14}.
The small size of these Wordnets (below 1000 synsets for high-F-score versions) underscores the difficulty of extracting sense information from unsupervised representations.
In particular, we found that stronger sense-induction methods, specifically sparse coding, than those presented in \cite{Nasiruddin:14} were needed to distinguish word-senses well.

Another approach is to leverage and expand upon existing resources.
Two multi-lingual Wordnets thus constructed are the Extended Open Multilingual Wordnet (OMW) \cite{omw}, which scraped Wiktionary, and the Universal Multilingual Wordnet (UWN) \citep{uwn}, which used multiple translations to match word-senses.
Through evaluation we found that the approach leads to high precision/low recall Wordnets.
This method is also used for BabelNet, which extends Wordnet and Wikipedia \citep{Navigli:12}.

Existing ontologies are also frequently used for sense representations;
these include efforts using Wordnet \citep{Rothe:15} and BabelNet \citep{Iacobacci:15}.
The approach often uses unsupervised embeddings for initialization and attains state-of-the-art on standard NLP tasks \citep{CamachoCollados:16}.
However, such representations depend on existing ontologies and so are difficult to apply to Wordnet construction.
We instead use unsupervised embeddings, shown empirically \citep{Mikolov:13} and under a generative model \citep{randwalk} to recover word-similarity and analogies from word-cooccurrences.
We use the latter paper's Squared-Norm (SN) vectors, which are similar in form to GloVe \citep{Pennington:14}.

\section{Distributed Synset Representation}
\label{sec:synsetrep}

\begin{figure*}[ht!]
\centering
\includegraphics[width=105mm]{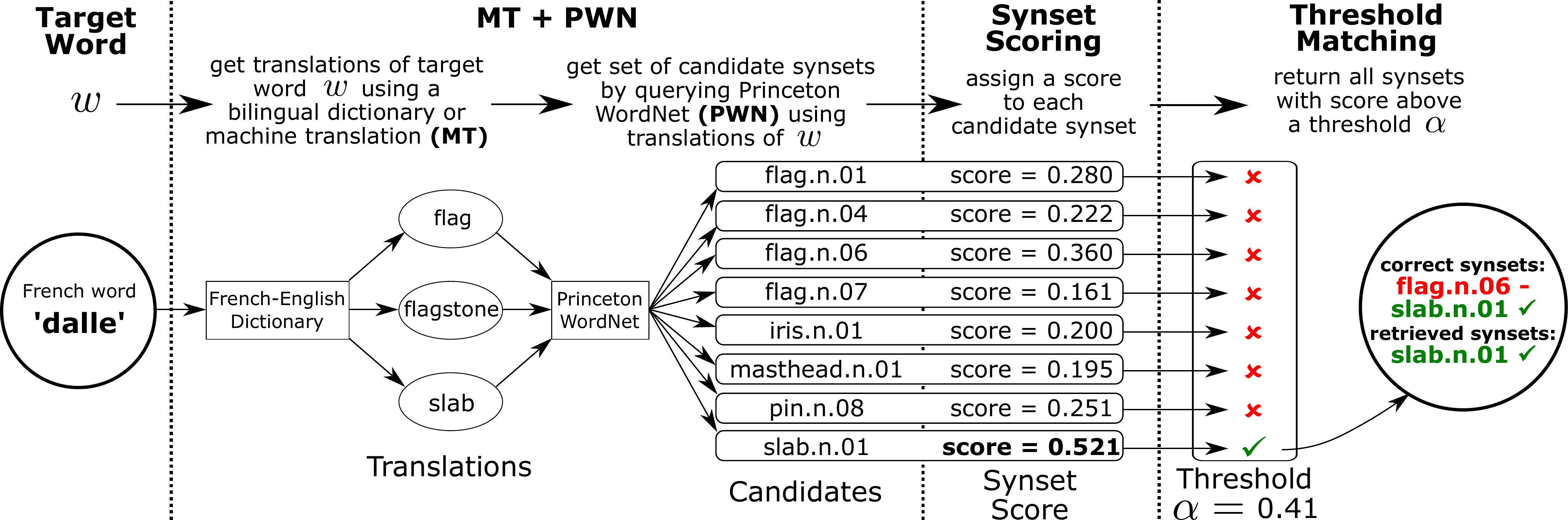}
\caption{
The score-threshold procedure for French word $w=$ {\em dalle} (flagstone, slab).
Candidate synsets generated by MT+PWN are given a score and matched to $w$ if the score is above a threshold $\alpha$.
}
\label{fig:threshold}
\end{figure*}

We introduce an unsupervised method for representing PWN synsets in non-English languages needing only a large corpus and machine translation.
For a vocabulary $V$ of target language words with $d$-dimensional\footnote{$d\ll|V|$, e.g. $d=300$ for vocabulary size $|V|=50000$.} unit vectors $v_w\in\mathbb{R}^d$, the representation of a synset $S$ will also be a vector $u_S\in\mathbb{R}^d$.
The construction of $u_S$ will be motivated by the following {\em score-threshold} procedure, illustrated in Figure \ref{fig:threshold}, for automated Wordnet construction.
Given a target word $w$, we use a bilingual dictionary to get its translations in English and let its set of candidate synsets be all PWN senses of the translations (MT+PWN).
We then assign a score $u_S\cdot v_w$ to each $S$ and accept as correct all synsets with score above a threshold $\alpha$;
if no synset is above the cutoff we accept only the highest-scoring synset.
Thus we want synset representations $u_S$ whose inner product with $v_w$ is high if $S$ is a matching synset of $w$ and low otherwise.
We present a simple baseline representation and then a more involved approach using embeddings of glosses.

\subsection{Baseline: Average Embedding}
\label{subsec:baseline}

Given a candidate synset $S$, define $T_S\subset V$ as the set of translations of its lemmas from English to the target language.
Then represent $S$ as
$$u_S=\frac{1}{|T_S|}\sum_{w\in T_S}v_w$$
In this case the synset score in the score-threshold procedure is equivalent to the average cosine similarity between $w$ and the translations of the lemmas of $S$.
Although straightforward, this representation is quite noisy and does not use all synset information provided by PWN.

\subsection{Synset Representation Method}
\label{subsec:synsetrep}

We now add synset relation and gloss information into the representation $u_S$.
Recalling the set $T_S$ of translations of lemmas of synset $S$, define $R_S$ to be the union over all synsets $S'$ related to $S$ of lemma-translation sets $T_{S'}$ .
Then the {\em lemma embedding} and {\em related-synset embedding} of $S$ are (before normalization) the element-wise sums 
$$v_{T_S}^{(SUM)}=\sum\limits_{w\in T_S}v_w\qquad\textrm{and}\qquad v_{R_S}^{(SUM)}=\sum\limits_{w\in R_S}v_w$$

While both gloss translations and the translated lemmas have mistakes from translation noise and polysemy, glosses also have irrelevant words (both stopwords and otherwise).
As we would like to downweight these, we use the sentence embedding formula of \cite{sentence}, a {\em smooth inverse frequency} (SIF) weighted average.
Given a list $L$ of words $w\in V$ with corpus frequency $f_w$, the SIF-embedding is
$$v_L^{(SIF)}=\sum\limits_{w\in L}\frac{a}{a+f_w}v_w$$
where $a$ is a parameter (commonly set to $10^{-4}$). Note that weight $\frac{a}{a+f_w}$ is low for high frequency words and so is similar to TF-IDF \citep{Salton:88}.
Through experiments with word-synset matching data, we found that simple sums work for representing the lemmas and related synsets of $S$ but SIF-embeddings are better for gloss representations.
Defining $D_S$ to be the translated synset definition of $S$ and $\mathcal{E}_S$ to be the set of translated example sentences of $S$, we set the {\em definition embedding} of $S$ to be $\hat{v}_{D_S}^{(SIF)}$ and the {\em example-sentence embedding} to be
$$\frac{1}{|\mathcal{E}_S|}\sum\limits_{E\in\mathcal{E}_S}\hat{v}_E^{(SIF)}$$
The representation $u_S$ of synset $S$ is then an average of all four (lemma, related-synset, definition, example-sentence\footnote{If $S$ has no example sentences this is not included in the average.}) of these embeddings.

\section{Cluster-Based Sense Representation}
\label{sec:senserep}

The representations above work well for automated Wordnets but make no use of the polysemous structure found to be encoded in embeddings themselves by \cite{polysemy}.
Here we describe their {\em Linear Word-Sense Induction} (Linear-WSI) model and introduce a {\em sense purification} procedure to represent each induced sense as a word-cluster.
Finally, we discuss an application to PWN sense clustering.
We again assume a vocabulary $V$ with each word $w$ represented by a unit vector $v_w\in\mathbb{R}^d$.

\subsection{Summary of Linear-WSI Model}
\label{subsec:wsi}

\cite{polysemy} posit that a vector of a word can be linearly decomposed into vectors associated to its senses.
Thus $w=$ {\em tie} --- which can be an article of clothing, a drawn match, and so on --- would be $v_w\approx av_{w\textrm{-clothing}}+bv_{w\textrm{-match}}+\dots$ for $a,b\in\mathbb{R}$.
Learning such fine-grained sense-vectors is difficult\footnote{Indeed this is a standard criticism of unsupervised approaches to WSI.}, but one expects some words to have related sense-vectors, e.g. the vector $v_{\textrm{tie-clothing}}$ would be close to the vector $v_{\textrm{bow-clothing}}$.
Thus Linear-WSI hypothesizes that for $k>d$ there exist unit basis vectors, or {\em atoms}, $a_1,\dots,a_k\in\mathbb{R}^d$ such that $\forall~w\in V$
\begin{equation}
\label{eqn:wsi}
v_w=\sum\limits_{i=1}^kR_{w,i}a_i+\eta_w
\end{equation}
where $\eta_w$ is a noise vector and at most $s$ coefficients $R_{w,i}$ are nonzero.
$v_w$ is thus modeled as a sparse linear combination of $s$ vectors $a_i$, with the hope that the sense-vectors $v_{\textrm{tie-clothing}}$ and $v_{\textrm{bow-clothing}}$ are both close to a clothing-related atom $a_i$.
(\ref{eqn:wsi}) is signals problem, {\em sparse coding}, that can be approximated by K-SVD \citep{Aharon:06}.
For $k=2000$ and $s=5$ \cite{polysemy} report that that the solution represents English word-senses as well as a competent non-native speaker and significantly better than clustering methods for WSI.

\subsection{Sense Purification}
\label{subsec:purification}

Though effective for WSI, the model produces comparatively few senses $a_i$ relative to the total number of synsets in WordNet;
indeed, if $k$ is set to be more than a few thousand the senses become repetitive.
For finer-grained representations we develop a {\em sense purification} procedure that views each sense as a pair $(w,a_i)$, where $a_i$ is a sense-vector s.t. $R_{w,i}>0$, and represents it as a cluster of words $C\subset V$.

For each word-sense pair $(w,a_i)$, sense purification finds a cluster $C$ of words whose embeddings are close to each other, to $v_w$, and to $a_i$.
The hope is that these words are used in contexts of $w$ in which the sense used is $a_i$.
Explicitly, given a word $w$, one of its senses $a_i$, and a fixed set-size $n$, we find $C$ as the $\argmax$ of:
\begin{align}
\label{eqn:objective}
\begin{split}
\maximize\limits_{C\subset V',C\ni w,|C|=n}&\qquad\gamma\\
\textrm{subject to}&\qquad\gamma\le\textrm{Median}\{v_x\cdot v_{w'}:w'\in C\backslash\{x\}\}~\forall~x\in C\\
&\qquad\gamma\le\textrm{Median}\{a_i\cdot v_{w'}:w'\in C\}
\end{split}
\end{align}
The constraints on the objective ensure that in order to maximize it the words $w'\in C$ must have high average cosine similarity with each other, with $w$, and with $a_i$.
For computational purposes we find $C$ approximately using a greedy algorithm that starts with $C=\{w\}$ and repeatedly adds to it the word $w\in V\backslash C$ that results in the highest objective value $\gamma$ of the new cluster.
Processing time is further reduced by restricting our search-space to be a subset of words in $V$ whose embeddings have cosine similarity of at least .2 with $v_w$ and $a_i$.

\begin{figure*}[ht!]
\centering
\includegraphics[width=50mm]{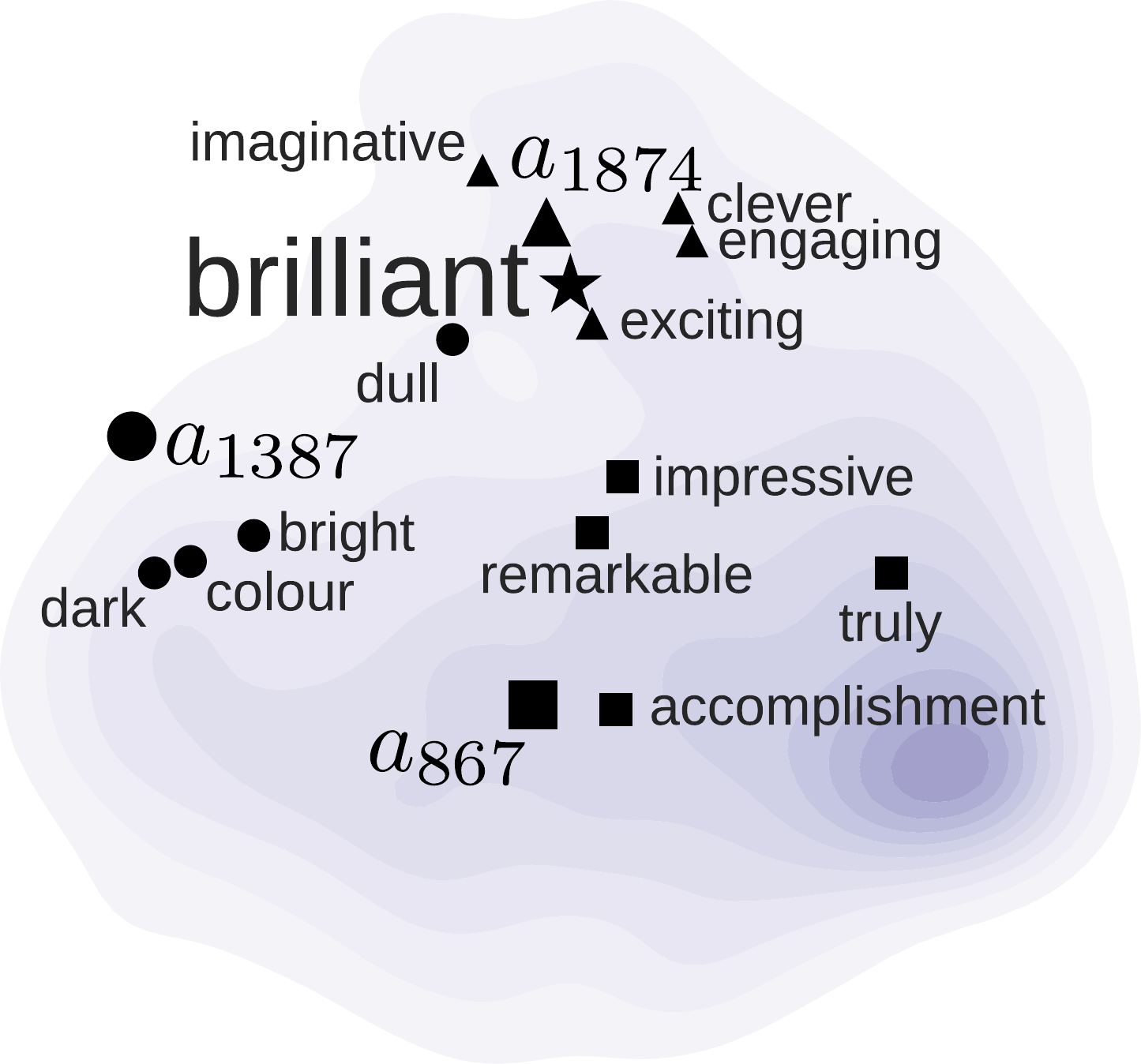}
\hspace{5mm}
\includegraphics[width=50mm]{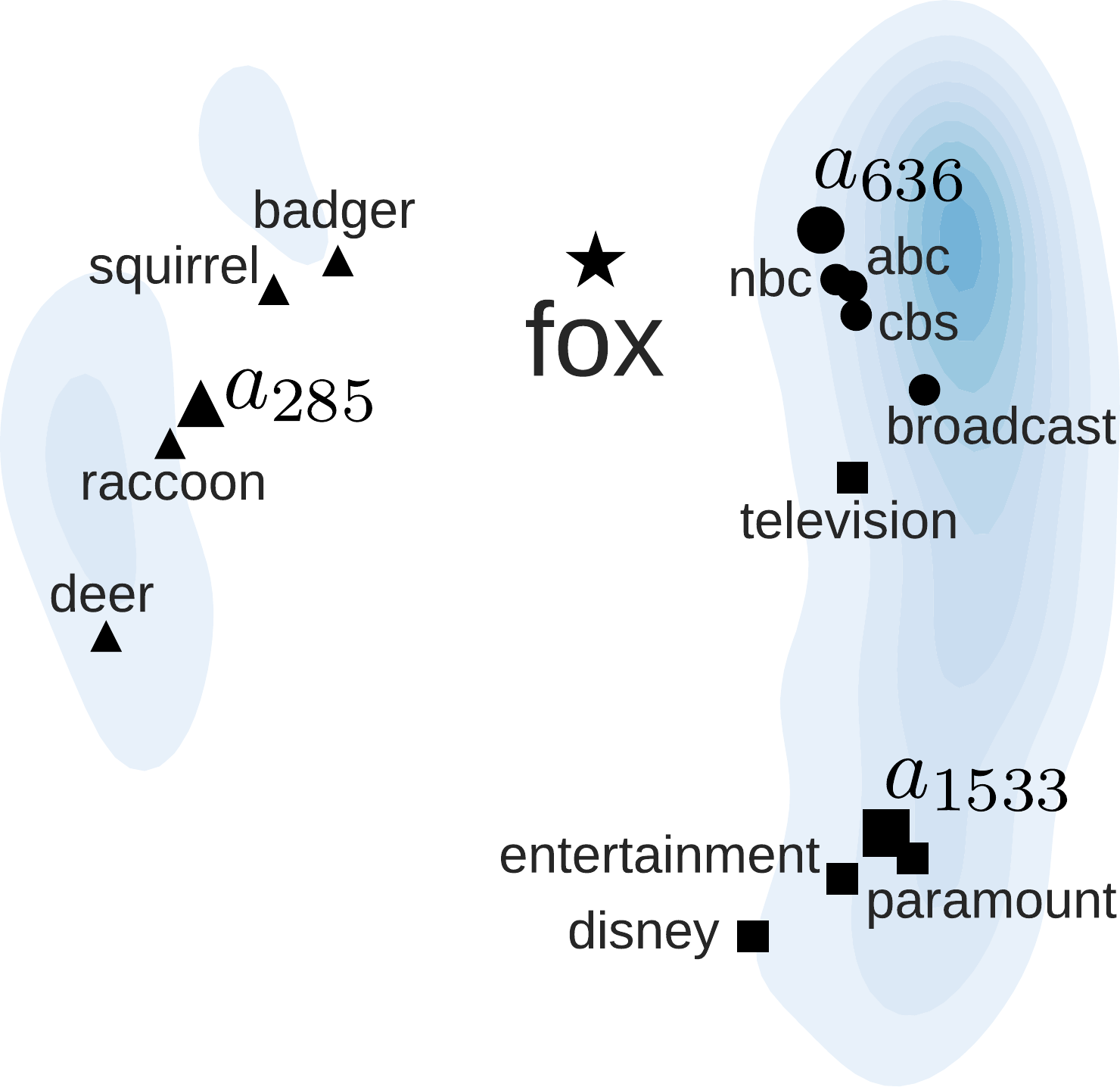}
\includegraphics[width=50mm]{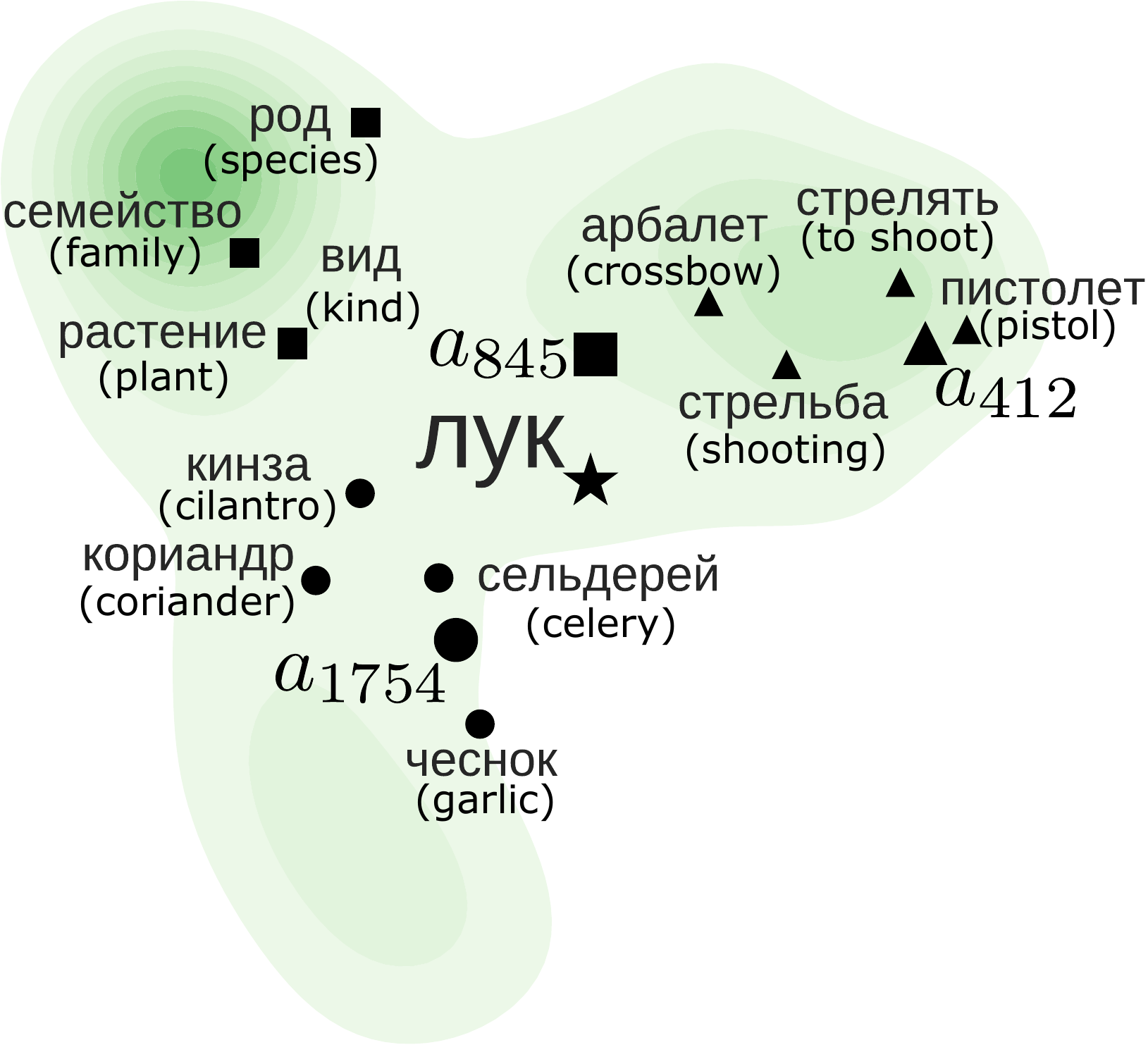}
\caption{
Isometric mapping of sense-cluster vectors for $w=$ {\em brilliant}, {\em fox}, and {\em\selectlanguage{russian}лук\selectlanguage{english}} (bow, onion). $w$ is marked by a star and each sense $a_i$ of $w$, shown by a large marker, has an associated cluster of words with the same marker shape. Contours are densities of vectors close to $w$ and at least one sense $a_i$. Note how correct senses are recovered across POS and languages and for both proper and common noun senses.}
\label{fig:isomap}
\end{figure*}

A depiction of the senses recovered via sense-purification is shown in Figure \ref{fig:isomap}.
Despite the difficulty of recovering small sense distinctions by distributional algorithms (partly due to Zipf's Law holding for word-senses), the algorithm is still able to distinguish very fine difference such as TV station Fox News vs. film corporation 20th Century Fox.

\subsection{Synset Clusters and Sense Clustering}
\label{subsec:clustering}

Thus far our work with word-senses has been entirely unsupervised, based only upon the polysemous structure of word embeddings.
We now consider an application of Linear-WSI and sense purification to the problem of {\em sense-clustering} --- reducing the granularity of Wordnet's sense distinctions by merging closely related senses of different words, This is a well-studied but difficult problem in NLP that is useful for applications requiring a much coarser set of senses for each word than that provided by PWN \citep{Agirre:03,Snow:07}.

To define our approach, we first specify a cluster similarity metric and a method for finding synset-atoms/synset-clusters for each word-synset pair $w,S$ using the atoms in the sparse representation of $v_w$.
This similarity condition (\ref{eqn:similar}) and the synset-atom/synset-cluster pairs $(a_S,C_S)$ will also be useful in improving Wordnet construction performance in the next section.

First, we take any two word-clusters $C_1,C_2\subset V$ and define a {\em cluster similarity function}
$$\rho(C_1,C_2)=\textrm{Median}\{v_x\cdot v_y:x\in C_1,y\in C_2\}$$
We then declare $C_1$ and $C_2$ to be {\em similar} if 
\begin{equation}
\label{eqn:similar}
\rho(C_1,C_2)\ge\min\{\rho(C_1,C_1),\rho(C_2,C_2)\}
\end{equation}
i.e. if their cluster similarity with each other exceeds either one's cluster similarity with itself.
Next, given a synset $S$ we define the set $V_S\subset V$ to be the union of all sets of translated lemmas of synsets related to $S$.
Then for any word-synset pair $w,S$ we let their {\em synset-atom} be the sense $a_i$ from all $a_i$ s.t. $R_{w,i}>0$ for which sense-purification using $V'=V_S$ as the search-space produces the {\em synset-cluster} $C_S$ with maximal objective value (see Equation \ref{eqn:objective}).
This can be done by running purification on each atom and choosing the best resulting cluster.

As formalized in Algorithm \ref{alg:clustering}, the sense-clustering algorithm merges synsets that share a sense $a_i$ in the sparse representation of $v_w$ and whose clusters share similar words.
Here the atoms $a_i$ s.t. $R_{w,i}>0$ represent the coarse set of senses of $w$ and each synset $S$ of $w$ is assumed to be related to one of them;
therefore merging synsets sharing an atom clusters those synsets together.

\begin{algorithm}[H]
\label{alg:clustering}
\SetAlgoLined
\SetAlgoNoEnd
\KwData{$w\in V$, its PWN synsets $\mathcal{S}$, and atoms $a_i$ s.t. $R_{w,i}>0$}
\For{candidate synset pairs $(S,S')\in\mathcal{S}\times\mathcal{S}$}{
compute synset-atoms $a_S,a_{S'}$ and synset-clusters $C_S,C_{S'}$\\
\If{$a_S=a_{S'}$ and $C_S,C_{S'}$ are similar (\ref{eqn:similar})}{merge the senses of $w$ associated with synsets $S$ and $S'$}}
\caption{Sense Clustering}
\end{algorithm}

\section{Methods for Automated Wordnet Construction}
\label{sec:wordnet}

Our basis for automated Wordnet construction is the score-threshold procedure described in Section \ref{sec:synsetrep}, where a candidate synset $S$ is matched to a word $w$ if $u_S\cdot v_w\ge\alpha$ for synset representation $u_S$ and a threshold $\alpha$.
The representation described in Section \ref{subsec:synsetrep} performs well compared to previous methods and our baseline;
however, through examination we identified two cases in which the method performs poorly: 
\begin{enumerate}
\item $w$ has no candidate synset $S$ with score $u_S\cdot v_w$ that clears the score-threshold $\alpha$.
\item $w$ has multiple closely related synsets that are all correct matches but some have a much lower score than the others.
\end{enumerate}
In this section we discuss how to improve performance in these cases by addressing a cause of noise in representing synset $S$ in the target language --- that due to polysemy many translated lemmas of $S$ and related synsets are irrelevant.
As seen before, sense-purification addresses a similar problem of Linear-WSI --- that each sense $a_i$ has too many related words --- by extracting a cluster of words related to both $w$ and $a_i$.
Thus synset clusters produced via purification as in Section \ref{subsec:clustering} may also lead to more useful representations of synsets than simply $u_S$.

Previously, given a word $w\in V$ we constructed a synset cluster $C_S$ and associated sense $a_S$ by using $V_S$, the union of the sets of all lemmas of synsets related to $S$, as the search-space $V'$ in sense-purification.
Since we now want synset clusters in the target vocabulary, we simply replace $V_S$ its translations.
Then for each candidate synset $S$ of $w$ we obtain an associated sense $a_S$ and cluster $C_S$ as in Section \ref{subsec:clustering}.

\begin{table*}[ht]
\centering
\begin{tabular}{@{}lll@{}}
Synset & $a_S$ & $C_S$ \\
\toprule
flag.n.01 & $a_{789}$ & poteau (goalpost), flèche (arrow), $\dots$ \\
flag.n.04 & $a_{892}$ & flamme (flame), fanion (pennant), $\dots$ \\
{\bf flag.n.06} & $a_{892}$ & dallage (paving), carrelage (tiling), $\dots$ \\
flag.n.07 & $a_{1556}$ & pan (section), empennage, queue, tail, $\dots$ \\
iris.n.01 & $a_{1556}$ & bœuf (beef), usine (factory), plante $\dots$ \\
masthead.n.01 & $a_{1556}$ & inscription, lettre (letter), $\dots$ \\
pin.n.08 & $a _{1556}$& trou (hole), tertre (mound), marais $\dots$ \\
{\bf slab.n.01} & $a_{892}$ & carrelage (tiling), carreau (tile) $\dots$ \\
\bottomrule
\end{tabular}
\caption{\label{tbl:clusters} Synset-atoms $a_S$ and clusters $C_S$ of {\em dalle} (flagstone, slab). The correct synsets (bold) have $C_S$ more related to their meaning.}
\end{table*}

\subsection{A Better Threshold Using the Purification Objective}
\label{subsec:better}

The first failure case of the score-threshold procedure --- no candidate synset scores above the cutoff $\alpha$ --- often occurs when synsets have little information in their glosses.
Letting $f(C):2^V\mapsto[0,1]$ be the objective function in Equation \ref{eqn:objective}, the synset-clusters $C_S$ obtained as above allow $f(C_S)$ to be used as another measure of relevance of $S$ with $w$, as an incorrect candidate synset $S$ likely has fewer translated related lemmas sharing a context with $w$ to put in the search-space $V_S$ for sense-purification and thus a lower objective value.

To exploit this, define $S^*=\arg\max f(C_S)$ as the synset whose cluster has maximal objective value.
Then replace $\alpha$ by a new cutoff $\alpha_w=\min\{\alpha,u_{S^*}\cdot v_w\}$ and match all candidate synsets $S$ with score $u_S\cdot v_w\ge\alpha_w$.
This ensures that if no synset's score is above $\alpha$, the synset $S^*$ with the best synset-cluster is matched to $w$ and, if it is polysemous, so are any candidate synsets $S$ of $w$ with score $u_S\cdot v_w\ge u_{S^*}\cdot v_w$.

\subsection{Recovering Similar Synsets Using Synset Clusters}
\label{subsec:recover}

The second failure case of the score-threshold procedure --- many similar candidate synsets of $w$ are correct but some have scores below the cutoff --- occurs for words with fine sense distinctions.
Thus we address the issue similarly to the sense clustering algorithm in Section \ref{subsec:clustering}.

\begin{algorithm}[H]
\label{alg:recover}
\SetAlgoLined
\SetAlgoNoEnd
\KwData{$w\in V$, its PWN synsets $\mathcal{S}$, and atoms $a_i$ s.t. $R_{w,i}>0$}
$\forall~S\in\mathcal{S}$ compute a synset-atom $a_S$ and a synset-cluster $C_S$\\
\For{each atom $a_i$}{
let $M_i\subset\mathcal{S}$ be candidates $S$ s.t. $a_S=a_i$ and score $u_S\cdot v_w\ge\alpha_w$\\
\For{each $S$ s.t. $a_S=a_i$ and $\beta\le u_S\cdot v_w<\alpha_w$}{
\If{$C_S$ and $C_{S'}$ are similar (\ref{eqn:similar}) $\forall~S'\in M_i$}{match synset $S$ to word $w$}}}
\caption{Synset Recovery}
\end{algorithm}

\begin{figure*}[ht!]
\centering
\includegraphics[width=105mm]{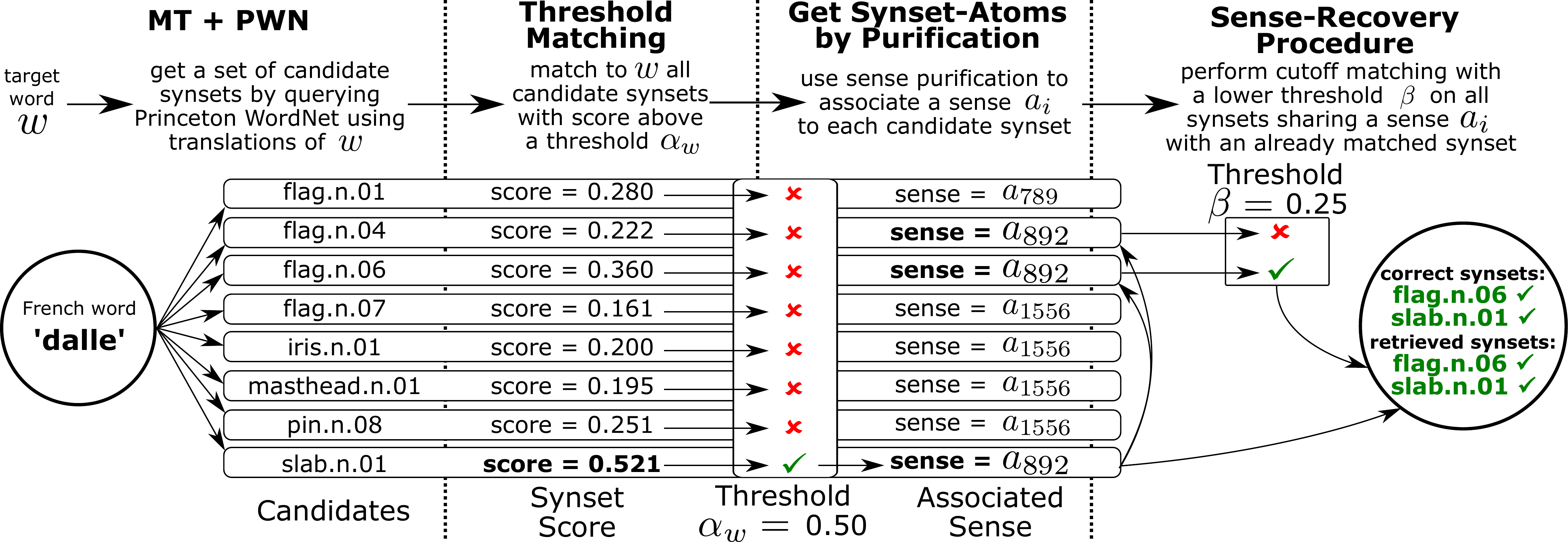}
\caption{
Score-threshold and sense-recovery procedures for French word $w=$ {\em dalle} (flagstone, slab).
Candidates are matched to $w$ if their score clears a cutoff $\alpha_w$.
If an unmatched synset shares a sense $a_i$ with a matched one, it is compared to a lower cutoff $\beta$ (sense-recovery).
}
\label{fig:wsi}
\end{figure*}

Given a word $w$, first run the score-threshold procedure with modified cutoff $\alpha_w$.
Then for a fixed low cutoff $\beta\le\alpha$ run Algorithm \ref{alg:recover}, allowing a candidate $S$ unmatched by the score-threshold procedure to be compared to $\beta\le\alpha$ if its synset-atom $a_S$ is the same as that of a matched synset $S'$ and their clusters $C_S,C_{S'}$ are similar.
This exploits the fact that similar synsets are likely associated with the same sense $a_i$ of $w$ and have similar words from which to construct their synset-clusters.
The final improvement of the score-threshold method with a $w$-dependent objective $\alpha_w$ and sense-recovery is outlined in Figure \ref{fig:wsi}.

\section{Evaluation of Automated Wordnet Construction}
\label{sec:results}

We evaluate our methods by constructing automated French and Russian Wordnets.
For word embeddings we train 300-dimensional SN vectors on $|V|\approx 50000$ words, restricted to those with 1000 occurrences or having candidate PWN synsets and 100 occurrences in the lemmatized Wikipedia corpus \citep{randwalk}.
We use sparsity $s=4$ and basis-size $k=2000$ for Linear-WSI and set-size $n=5$ for sense-purification. 
Translation dictionaries were built from Google and Microsoft Translate and the dictionary of the translation company ECTACO;
Microsoft was used for the sentence-length MT needed for translating synset glosses.

\subsection{Test Sets for Word-Synset Matching}
\label{subsec:testsets}

A natural method of evaluation is by using a manually constructed Wordnet as a source of ``ground truth" senses.
However, the ELRA French Wordnet\footnote{\url{http://catalog.elra.info/product_info.php?products_id=550}} is private while Russian Wordnets are too small and unlinked with PWN\footnote{\url{http://project.phil.spbu.ru/RussNet/}} or gotten by direct translation from PWN\footnote{\url{http://wordnet.ru/}}.

We instead construct and release test sets for each language by randomly choosing 200 each of adjectives, nouns, and verbs from words whose English translations appear in the synsets of the {\em Core WordNet}, a semi-automatically selected set of about 5000 most-used synsets in PWN \citep{Fellbaum:98}. 
Choosing words from Core synset lemmas makes the evaluation more difficult since common words are more polysemous, with more synsets to retrieve;
this is reflected in the lower performance of WOLF relative to \citep[Table 4]{wolf}.

The ``ground truth" senses are picked by native speakers asked to match synsets to a word given a set of candidates synsets generated by MT+PWN.
For example, the French word {\em foie} has one translation, {\em liver}, with four PWN synsets: 
1-``glandular organ"; 2-``liver used as meat"; 3-``person with a special life style"; 4-``someone living in a place." 
Only the first two align with senses of {\em foie}, so the expert marks the first two as good and the others as negative. 

Two native speakers of each language were trained by a conversant author with knowledge of WordNet;
the latter also resolved discrepancies.
We get 600 words and $\sim 12000$ candidate word-synset pairs in each language, with adjectives and nouns having on average about 15 candidates and verbs having about 30.
This comprises a very large data set compared to previous efforts.
Accuracy compared to this ground truth estimates how well an algorithm does compared to humans.

One property of this test set is its dependence on the translation we use to get candidate synsets, which can leave out correct synset matches if they are not in the bilingual dictionaries.
However, providing both correct and incorrect candidates allows future work to focus on selecting senses and not worry about finding the best dictionary.
This dictionary-independent evaluation is an important feature since translation systems used by many authors are often not provided in full.
When comparing our performance to previous work, we do not penalize word-synset matches in which the synset is not among the candidates generated for that word, reducing the loss of precision incurred by other methods due to the use of different dictionaries. 
We also do not penalize other Wordnets for test words they do not contain.

In addition to precision and recall, we report {\em coverage} as the proportion of synsets in the Core WordNet that are matched to.
While an imperfect metric given different sense usage by language, the synsets are universal-enough for it to be a good indicator of usability.

\subsection{Experimental Results}
\label{subsec:results}

\begin{table*}[ht]
\centering
\begin{threeparttable}
\fontsize{7pt}{8.4pt}\selectfont
\begin{tabular}{@{}cccccccr@{}}
Method & POS & $F_{.5}^\ast$ & $\textrm{Prec.}^\ast$ & $\textrm{Rec.}^\ast$ & Coverage & Synsets \\
\toprule
\multirow{4}*{\specialcell{Wordnet Libre\\du Fran\c{c}ais\\(WOLF)\\{\tiny \citep{wolf}}}} 
& Adj. & 66.3 & 78.1 & 53.4 & 84.8 & 6865 \\
& Noun & 68.6 & 83.2 & 51.5 & 95.0 & 36667 \\
& Verb & 60.8 & 81.0 & 39.6 & 88.2 & 7671 \\
& Total & 65.2 & 80.8 & 48.2 & 92.2 & $\textrm{52757}^\dagger$ \\
\midrule
\multirow{4}*{\specialcell{Universal Wordnet\\(UWN)\\{\tiny \citep{uwn}}}} 
& Adj. & 64.5 & 88.3 & 42.3 & 69.2 & 7407 \\
& Noun & 67.5 & 94.1 & 40.8 & 75.9 & 24670 \\
& Verb & 55.4 & 88.0 & 28.5 & 76.2 & 5624 \\
& Total & 62.5 & 90.1 & 37.2 & 75.0 & $\textrm{39497}^\dagger$ \\
\midrule
\multirow{4}*{\specialcell{Extended Open\\Multilingual Wordnet\\(OMW)\\{\tiny \citep{omw}}}} 
& Adj. & 58.4 & {\bf 90.9} & 28.4 & 54.7 & 2689 \\
& Noun & 61.3 & {\bf 96.5} & 31.7 & 66.6 & 14936 \\
& Verb & 47.8 & {\bf 95.9} & 18.6 & 57.7 & 2331 \\
& Total & 55.9& {\bf 94.5} & 26.2 & 63.2 & $\textrm{20449}^\dagger$ \\
\midrule
\multirow{4}*{\specialcell{Baseline:\\Average Similarity\\(Section~\ref{subsec:baseline})}} 
& Adj. & 62.8 & 65.3 & {\bf 68.5} & 88.7 & 9687 \\
& Noun & 67.3 & 71.6 & {\bf 69.0} & 92.2 & 37970 \\
& Verb & 51.8 & 55.9 & {\bf 57.0} & 83.5 & 10037 \\
& Total & 60.6 & 64.3 & {\bf 64.9} & 90.0 & $\textrm{58962}^\dagger$ \\
\midrule
\multirow{4}*{\specialcell{Method 1:\\Synset Representation\\(Section~\ref{subsec:synsetrep})}} 
& Adj. & 65.9 & 75.9 & 59.5 & 85.1 & 8512 \\
& Noun & 71.0 & 78.7 & 69.1 & \bf 96.7 & 35663 \\
& Verb & 61.6 & 78.7 & 49.8 & 89.9 & 8619 \\
& Total & 66.2 & 77.8 & 59.5 & \bf 93.7 & $\textrm{53852}^\dagger$ \\
\midrule
\multirow{4}*{\specialcell{Method 2:\\Synset Representation\\+ Linear-WSI\\(Section~\ref{sec:wordnet})}} 
& Adj. & {\bf 67.7} & 76.9 & 62.6 & \bf 91.2 & 8912 \\
& Noun & {\bf 73.0} & 83.7 & 62.0 & 90.9 & 34001 \\
& Verb & {\bf 64.4} & 79.3 & 51.5 & \bf 93.6 & 9262 \\
& Total & {\bf 68.4} & 80.0 & 58.7 & 91.5 & $\textrm{53208}^\dagger$ \\
\bottomrule
\end{tabular}
\begin{tablenotes}
\item[$\ast$] Parameters tuned on a random-selected half of the data; evaluation done on the other half. All percentages are accurate within .2 with 95\% confidence.
\item[$\dagger$] Includes adverb synsets using same parameter values ($\alpha$ and $\beta$) as for adjectives.
\end{tablenotes}
\end{threeparttable}
\caption{\label{tbl:fr}French Wordnet Results}
\end{table*}

We report results in Tables~\ref{tbl:fr}~\&~\ref{tbl:ru}.
Parameters $\alpha$ and $\beta$ are tuned to maximize micro-average $F_{.5}$-score $\frac{1.25\cdot\textrm{Precision}\cdot\textrm{Recall}}{.25\cdot\textrm{Precision}+\textrm{Recall}}$, used instead of $F_1$ to prioritize precision (often more important for applications).
Our synset representations (Section~\ref{subsec:synsetrep}) outperform the baseline by 6\% in $F_{.5}$-score for French and 10\% for Russian;
in French it is competitive with (WOLF) and in both it exceeds both multi-lingual Wordnets.
Linear-WSI heuristics further improve $F_{.5}$-score by 1\% in Russian and 2\% for French, exceeding WOLF in $F_{.5}$-score across POS while having similar coverage.
Notably, OMW consistently achieves best precision, although it and UWN have low recall and coverage.

\begin{table*}[ht]
\centering
\begin{threeparttable}
\fontsize{7pt}{8.4pt}\selectfont
\begin{tabular}{@{}ccccccccrcc@{}}
Method & POS & $F_{.5}^\ast$ & $\textrm{Prec.}^\ast$ & $\textrm{Rec.}^\ast$ & Coverage & Synsets \\
\toprule
\multirow{4}*{\specialcell{Universal Wordnet\\(UWN)\\{\tiny \citep{uwn}}}} 
& Adj. & 52.4 & 80.3 & 29.6 & 51.0 & 11412 \\
& Noun & 65.0 & 87.5 & 45.1 & 71.1 & 19564 \\
& Verb & 48.1 & 74.8 & 25.7 & 65.0 & 3981 \\
& Total & 55.1 & 80.8 & 33.4 & 67.1 & $\textrm{30015}^\dagger$ \\
\midrule
\multirow{4}*{\specialcell{Extended Open\\Multilingual Wordnet\\(OMW)\\{\tiny \citep{omw}}}} 
& Adj. & 58.7 & {\bf 91.7} & 29.2 & 55.3 & 2419 \\
& Noun & 67.8 & {\bf 93.5} & 42.5 & 68.4 & 14968 \\
& Verb & 51.1 & {\bf 84.5} & 23.9 & 56.6 & 2218 \\
& Total & 59.2 & {\bf 89.9} & 31.9 & 64.2 & $\textrm{19983}^\dagger$ \\
\midrule
\multirow{4}*{\specialcell{Baseline:\\Average Similarity\\(Section~\ref{subsec:baseline})}} 
& Adj. & 61.4 & 60.9 & {\bf 77.3} & 92.1 & 10293 \\
& Noun & 55.9 & 59.9 & 59.9 & 77.0 & 32919 \\
& Verb & 46.3 & 49.0 & {\bf 55.1} & 84.1 & 9749 \\
& Total & 54.5 & 56.6 & {\bf 64.1} & 80.5 & $\textrm{54372}^\dagger$ \\
\midrule
\multirow{4}*{\specialcell{Method 1:\\Synset Representation\\(Section~\ref{subsec:synsetrep})}} 
& Adj. & 69.5 & 78.1 & 61.7 & 84.2 & 8393 \\
& Noun & 69.8 & 77.6 & 66.0 & 85.2 & 29076 \\
& Verb & 54.2 & 63.3 & 57.4 & 91.2 & 8303 \\
& Total & 64.5 & 73.0 & 61.7 & 86.3 & $\textrm{46911}^\dagger$ \\
\midrule
\multirow{4}*{\specialcell{Method 2:\\Synset Representation\\+ Linear-WSI\\(Section~\ref{sec:wordnet})}} 
& Adj. & {\bf 69.7} & 77.3 & 63.6 & \bf 93.3 & 9359 \\
& Noun & {\bf 71.6} & 78.1 & {\bf 68.0} & \bf 91.0 & 31699 \\
& Verb & {\bf 54.4} & 64.9 & 52.6 & \bf 91.9 & 8582 \\
& Total & {\bf 65.2} & 73.4 & 61.4 & \bf 91.5 & $\textrm{50850}^\dagger$ \\
\bottomrule
\end{tabular}
\begin{tablenotes}
\item[$\ast$] Parameters tuned on a random-selected half of the data; evaluation done on the other half. All percentages are accurate within .2 with 95\% confidence.
\item[$\dagger$] Includes adverb synsets using same parameter values ($\alpha$ and $\beta$) as for adjectives.
\end{tablenotes}
\end{threeparttable}
\caption{\label{tbl:ru}Russian Wordnet Results}
\end{table*}

Across POS, we do best on nouns and worst on verbs, a standard result likely exacerbated in this case due to the greater polysemy of verbs.
Comparing between languages, we see slightly better performance on Russian adjectives, slightly worse performance on Russian nouns, and much worse performance on Russian verbs.
The latter can be explained by a difference in treating the reflexive case and aspectual variants due to the grammatical complexity of Russian verbs.
In French, making a verb reflexive requires adding a word while in Russian the verb itself changes, e.g. {\em to wash}$\to${\em to wash oneself} is {\em laver}$\to${\em se laver} in French but {\em\selectlanguage{russian}мыть\selectlanguage{english}}$\to${\em\selectlanguage{russian}мыться\selectlanguage{english}} in Russian. 
Thus we do not distinguish them for French as the token is the same but for Russian we do, so both {\em\selectlanguage{russian}мыть\selectlanguage{english}} and {\em\selectlanguage{russian}мыться\selectlanguage{english}} may appear and have distinct synset matches. 
Matching Russian verbs is thus harder as the reflexive usage is often contextually similar to the non-reflexive usage. 
Aspectual verb pairs are another complication; for Russian, {\em to do} has aspects {\em\selectlanguage{russian}(делать, сделать)\selectlanguage{english}} that are treated as distinct while in French these are just tenses of {\em faire}.

Overall the word embedding method seems robust to the language's closeness to English, with similar noun and adjective performance and a verb-performance discrepancy stemming from an intrinsic quality rather than language dissimilarity.
Such a claim can be further examined by constructing Wordnets for non-European languages.

\section{Conclusion}

We have introduced unsupervised synset and sense representations via word vectors that can be used to improve WordNet and extend it to other languages.
These methods outperform language-specific and resource-heavy approaches, enabling the construction of automated Wordnets in low-resource languages.
We also release two large POS-split test sets for automated Wordnets for French and Russian that give a more accurate picture of a method's strengths and weaknesses.
In future work these methods may be improved upon by incorporating other language representation methods such as multi-lingual embeddings \citep{Faruqui:14}.
Furthermore, the sense-purification procedure we introduce has direct applications to word-sense induction, clustering, and disambiguation.

\bibliographystyle {cslipubs-natbib}
\bibliography {lilt2017}

\end{document}